\documentclass[conference]{IEEEtran}
\IEEEoverridecommandlockouts

\usepackage{cite}
\usepackage{amsmath,amssymb,amsfonts}
\usepackage{graphicx}
\usepackage{xcolor}
\usepackage{booktabs}
\usepackage{array}
\usepackage{listings}
\usepackage{algorithm}
\usepackage{algpseudocode}
\usepackage{tikz}
\usetikzlibrary{arrows.meta,positioning,shapes.geometric}
\usepackage[hidelinks]{hyperref}
\usepackage{url}
\usepackage{textcomp}
\usepackage{cuted}
\usepackage{capt-of}
\newcommand{\runin}[1]{\par\smallskip\noindent\textbf{#1.}\hspace{0.5em}}

\newcommand{\file}[1]{\path{#1}}

\tikzset{
  pbox/.style={draw,rounded corners=1.4pt,align=center,font=\scriptsize,
               inner sep=3pt,minimum height=6.5mm,line width=0.4pt},
  pdec/.style={pbox,fill=black!4},
  pflow/.style={-{Latex[length=1.6mm,width=1.2mm]},line width=0.4pt},
}

\begin{document}

\title{WetRobo: A Reproducible Robot Kit for\\Coding Agents in Biological Laboratories}
\hypersetup{
  pdftitle={WetRobo: A Reproducible Robot Kit for Coding Agents in Biological Laboratories},
  pdfauthor={Yuna Oikawa, Kei Endo, Takanori Uzawa, Yunzhe Zhang, Manan Anjaria, Lerrel Pinto, Sherry Yang, Koji Tsuda}
}

\author{
\IEEEauthorblockN{Yuna Oikawa\textsuperscript{1}, Kei Endo\textsuperscript{1},
Takanori Uzawa\textsuperscript{2}, Yunzhe Zhang\textsuperscript{3},\\
Manan Anjaria\textsuperscript{3}, Lerrel Pinto\textsuperscript{3},
Sherry Yang\textsuperscript{3,4}, Koji Tsuda\textsuperscript{1,5,6}}
\IEEEauthorblockA{\textsuperscript{1}Department of Computational Biology and Medical Sciences,\\
The University of Tokyo, Japan\\
\textsuperscript{2}RIKEN Pioneering Research Institute, Japan\qquad \textsuperscript{3}New York University, USA\\
\textsuperscript{4}Google DeepMind, USA\\
\textsuperscript{5}National Institute for Materials Science, Japan\qquad \textsuperscript{6}RIKEN Center for Advanced Intelligence Project, Japan\\
\href{mailto:tsuda@k.u-tokyo.ac.jp}{tsuda@k.u-tokyo.ac.jp}}
}

\maketitle

\begin{abstract}
Automating biological research requires general-purpose, reproducible robot
systems that allow individual wet-lab researchers to delegate robot tasks
without performing teleoperation or neural-network training.
Vision-language-action policies have been proposed for general-purpose arms,
but can lose performance when their operating environment changes. We
therefore built WetRobo, a robot kit that can readily transfer between
laboratories. It consists of one robot
arm, laboratory equipment (an incubator, a reagent bottle with a cap, and a Petri
dish), the existing code that moves the arm, teleoperation demonstrations of
each task that we recorded, and a general \texttt{AGENTS.md} skill file. A biological experimentalist
provides natural-language tasks without collecting local teleoperation
training data or training a neural network. The coding agent observes the local laboratory and writes
and executes programs, using external tools as needed for adaptation. We
demonstrate use of WetRobo with OpenAI Codex (\texttt{gpt-5.6-sol}) on
three successful tasks: lifting a Petri dish lid, removing a bottle cap, and
opening the incubator door, all in real-world laboratories. The coding agent achieved the cap task in both
laboratories, Lab X and Lab Y, whereas
a VLA fine-tuned on Lab X demonstrations succeeded there but failed to
transfer to Lab Y. These results point to a practical route for laboratory
robotics: instead of training a policy for each laboratory, distribute a
kit and let a coding agent adapt it in each laboratory.
Code, demonstrations, and the evolved programs are available at
\url{https://github.com/tsudalab/WetRobo}.
\end{abstract}

\begin{strip}
\centering
\includegraphics[width=\textwidth]{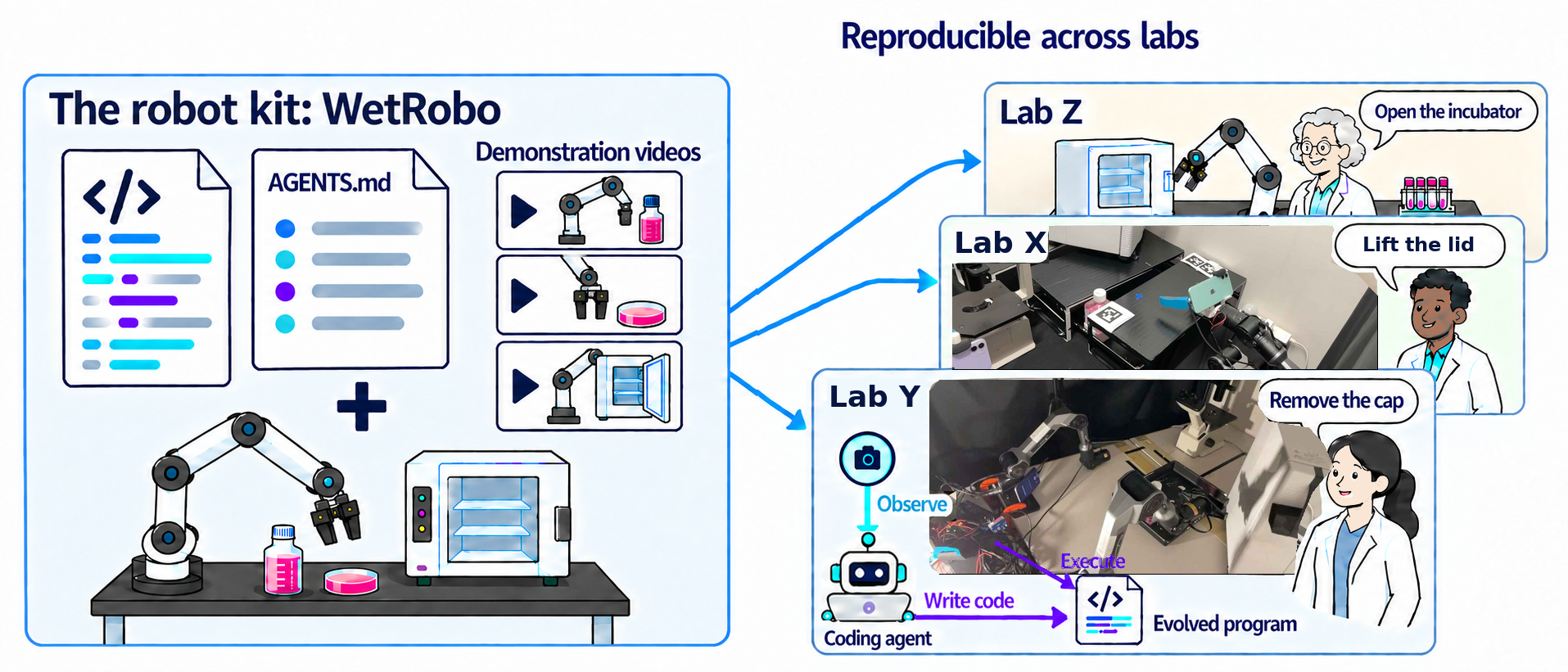}
\captionof{figure}{Deployment of WetRobo. The kit combines
\texttt{AGENTS.md}, the base stack, and demonstration videos with a robot
arm and laboratory equipment. Biological experimentalists in laboratories including
Lab X and Lab Y, assemble the equipment and provide natural-language tasks without
collecting teleoperation data or training neural networks; a coding agent observes each local setting,
writes an evolved program, and executes it on the robot.}
\label{fig:concept}
\end{strip}

\section{Introduction}\label{introduction}

Biological experimentation rests on a large amount of repetitive manual
labor. In cell culture, cells must be fed, passaged, and inspected at
intervals set by their growth. While large, centralized factory-laboratories already have automation
options ranging from robotic high-throughput screening \cite{qhts}
to cloud laboratories \cite{cloudlab}, individual laboratories need more affordable and general-purpose strategies
that accommodate their existing equipment. RoboCulture provides a flexible
robotic platform for liquid handling and cell culture \cite{roboculture}.
More general-purpose laboratory robots include Maholo, whose dual-arm
manipulation was combined with Bayesian optimization to improve a cell
differentiation protocol \cite{maholo}.

Vision-language-action (VLA) models, including $\pi_{0.5}$ \cite{pi05},
offer a general-purpose policy interface increasingly studied for laboratory
manipulation. Pipette \cite{pipette} and AutoBio \cite{autobio} train and
evaluate VLAs on simulated wet-lab tasks. LabVLA introduces laboratory-specific
training across robot embodiments \cite{labvla}, while BioProVLA-Agent
combines protocol reasoning and visual verification with VLA policies
fine-tuned for biological manipulation \cite{bioprovla}. ProtoAct links
protocols to actions and supports demonstration collection and VLA training
\cite{protoact}. However, robustness evaluations of VLA policies reveal strong
sensitivity to changes in camera viewpoints and robot initial states
\cite{liberoplus}. Real-world experiments on a Franka arm report sharp
degradation of a frozen VLA under appearance, position, and object changes,
including failure on an unseen replacement object \cite{vls}.

Code-as-policy offers a route to addressing this adaptation challenge through
executable-code interfaces for robot control and task
planning \cite{codeaspolicies, progprompt}.
Rather than directly predicting actions with an imitation-trained policy,
a language model writes executable code that generates actions.
Task descriptions and observations inform its reasoning, expressed as
inspectable, editable program logic \cite{codeaspolicies, progprompt}.
CaP-X is an open-access framework for systematically studying code-as-policy
agents in robot manipulation \cite{capx}.
Recently, ENPIRE was introduced as a harness framework for coding agents to
autonomously control robots and improve their policies through feedback
loops \cite{enpire}. The authors demonstrated coding agents' ability to perform
real-world tasks including GPU insertion, pin insertion, Push-T, and zip tie
cutting.
We bring this approach to biological laboratory automation by creating
a domain-specific \texttt{AGENTS.md}, adding demonstrations as references
for the robot, and specifying laboratory hardware checked for automation;
we distribute these components together as a robot kit.

WetRobo (Fig.~\ref{fig:concept}) supplies biological experimentalists with arm
and equipment specifications, the base stack, and a skill file with operating
rules and a starting workflow. Whereas ENPIRE establishes reset and verification
interfaces through human-guided setup \cite{enpire}, WetRobo leaves task programs
and success checks to on-site adaptation. Its \texttt{AGENTS.md} supplies handling
rules for culture equipment and the incubator while encouraging the agent to
select external tools and define success criteria from local observations.
The kit also includes fifteen teleoperated
demonstrations per task.
The agent can inspect or replay them to understand the demonstrated handling.
Biological experimentalists assemble the equipment and give natural-language
tasks; the agent handles adaptation, replacing local teleoperation data
collection and neural-network training. Observation and code execution leave
an evolved program and an action record.

To demonstrate WetRobo's capability, we gave it to OpenAI Codex \cite{codex}
for three tasks, achieving Petri-lid lifting, bottle-cap removal, and
incubator-door opening on physical equipment in laboratories. In this evaluation of code-as-policy for real wet-lab
scientific work, the coding agent succeeded at bottle-cap lifting in two independent
laboratories, whereas a VLA fine-tuned on Lab X demonstrations succeeded in
Lab X but failed to transfer to Lab Y. The contributions are:
(i) the development of WetRobo, a kit that lets biological experimentalists
delegate laboratory tasks to a coding agent without training a neural
network or collecting teleoperation data themselves;
(ii) a demonstration that a coding agent evolves its programs and adapts to
a laboratory setting, accomplishing Petri-lid lifting, bottle-cap removal,
and incubator-door opening;
(iii) evidence that WetRobo with a coding agent transfers task capability
across laboratories more effectively than the evaluated VLA in the cap task,
suggesting a practical direction for laboratory robotics.

\section{The Input: WetRobo}\label{sec:wetrobo}

WetRobo is a combination of a git repository, \texttt{wetrobo}, a Piper arm,
example laboratory equipment, and 15 demonstrations per task. The repository
contains essentially the skill file \texttt{AGENTS.md} and
the basic code that drives the Piper arm. We call that code, as shipped, the
\emph{base stack} (branch \texttt{main}), and the code the agent leaves after a trial the
\emph{evolved program}; trial snapshots remain as branches of this same repository.
The repository is available at \url{https://github.com/tsudalab/WetRobo}.

\noindent\begin{minipage}{\columnwidth}
\subsection{\texttt{AGENTS.md}}\label{sec:skills}

\centering
\includegraphics[width=\columnwidth]{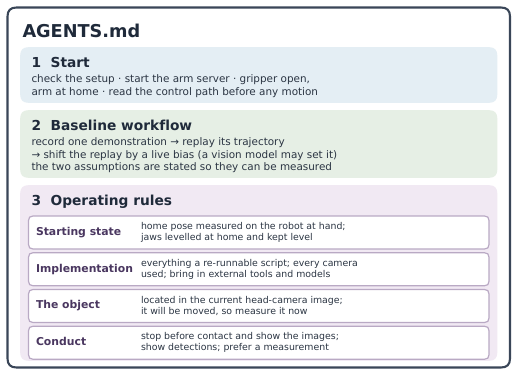}
\captionof{figure}{Structure of \texttt{AGENTS.md}: setup, demonstration replay with
live bias, and operating rules grouped by purpose.}
\label{fig:skills}
\end{minipage}\par\medskip

The file contains setup, a baseline workflow of demonstration replay with
live bias, and operating rules on starting state, implementation, the object,
and conduct (Fig.~\ref{fig:skills}).
It explicitly encourages external Python packages, models, and methods
available online, beyond tools already imported by the repository.

For work on the incubator and bench-top culture equipment, the rules require
horizontal jaws through approach, grasp, and transport unless a tilt is
needed---a constraint suited to open containers with contents.
Verification takes priority over speed: motion pauses before contact for
image-based pose confirmation, while configured keep-out zones and per-step
motion caps provide preventive geometric checks, with a stop available
because collision reactions are absent.
Home poses must be measured locally and images checked for shadows under
changed lighting rather than relying on assumptions from another laboratory.
Together, these rules prescribe motion suited to biological experimentation.

\subsection{Piper arms}\label{sec:arm}

The robot kit contains two six-degree-of-freedom AgileX Piper arms on CAN.
Every execution reported here uses the right arm.

\subsection{Cameras}\label{sec:cameras}

The kit uses head and wrist cameras, with camera counts varying between
laboratories. \file{robot/camera_map.json} maps head, left, and right roles
to device indices. iPhones supply RGB-D observations through Record3D
\cite{record3d}; base-stack capture saves RGB images, depth, and camera
intrinsics in run directories.

\subsection{Base stack}\label{sec:code}

The base stack code in the \texttt{wetrobo} repository drives the arm and
the laboratory's cameras. It contains an arm RPC server
(\file{robot/cone_e.py}) with a workspace clamp, and replay of a recorded
HDF5 demonstration through \file{rollout/controller.py}. In the replay a
per-arm Cartesian bias is added, and a \texttt{SafetyLayer} applies keep-out
zones and a per-step motion cap. It also supports live bias setting.
It carries no evolved program.

\subsection{Laboratory equipment}\label{sec:equipment}

The tasks act on three objects at the bench in front of the arm: a laboratory
incubator (EYELA LTI-300), whose door handle is a recessed horizontal slot;
a 500\,mL reagent bottle (Nacalai Tesque D-PBS($-$), $1\times$) with a small
white screw cap; and a Petri dish with a lid (TPP Techno Plastic Products AG 93100). The kit specifies equipment
physically tested with the arm and gripper, allowing laboratories that install the kit
to assemble a matching starting setup. The verified operations comprise
bottle-cap removal, incubator-door opening, and Petri-lid lifting.

\subsection{Demonstrations}\label{sec:demonstrations}

The kit includes 15 teleoperation demonstrations per task. They were
recorded with a Meta Quest teleoperation interface. The replay path in
Section~\ref{sec:code} can execute any of them.
For example, the saved incubator-door reference encodes demonstrated
approach, grasp, and opening motions for that appliance.
Demonstrations are optional references for the coding agent to consult when
useful, rather than prerequisites for execution. The agent removed
the reagent bottle cap without consulting the task demonstrations
(Section~\ref{sec:cap}).

\section{Results: Evolution of the Programs}\label{sec:results}

\begin{figure}[!h]
\centering
\includegraphics[width=\columnwidth]{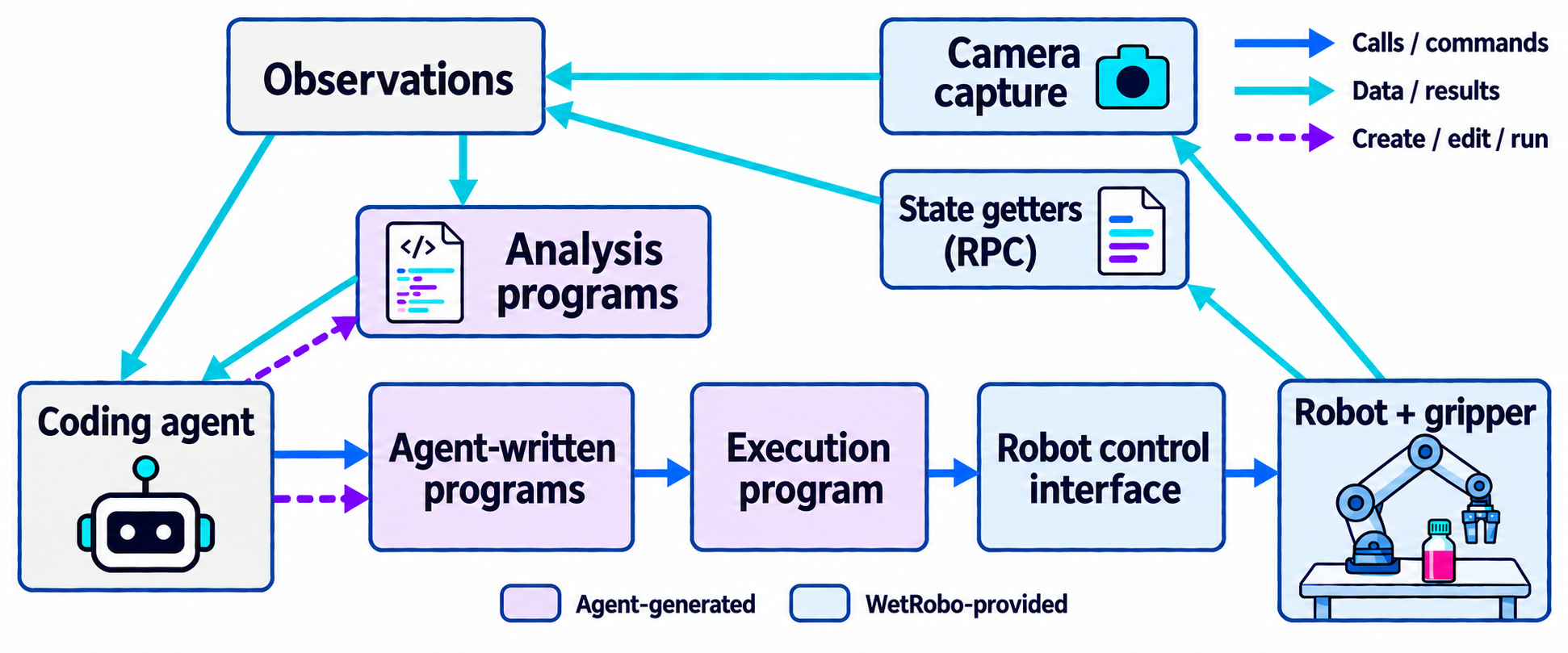}
\caption{How the coding agent observes and controls the robot during task performance.}
\label{fig:agent-control-flow}
\end{figure}

Fig.~\ref{fig:agent-control-flow} describes how observation leads to action.
WetRobo supplies the arm, gripper, cameras, state readings, and command interfaces.
The agent interprets images, joint angles, and gripper opening directly or
generates analysis programs to measure positions, alignment, and progress.
The agent also writes motion/check coordination and execution programs.
Execution parameters include movement
distances and directions, target positions and orientations, rotation angles,
gripper opening, the size of each motion step, and limits for checks.
An execution program converts a requested hand movement into joint movements,
or sends a sequence of hand positions and orientations. WetRobo's control
interface delivers these and gripper commands through RPC calls to the robot.
Images, opening, joint state, and available torque or effort check progress
and contact.

WetRobo was given to the OpenAI Codex CLI running
\texttt{gpt-5.6-sol} with reasoning effort \emph{medium} and approval policy
\emph{never}. Tasks were given as prompts.
Each trial left evolved programs and issued actions; we examined these and
the chat's reasoning, and we report them here.

\subsection{Petri dish lid}\label{petri-dish-lid}

\begin{figure}[!h]
\centering
\includegraphics[width=0.8\columnwidth]{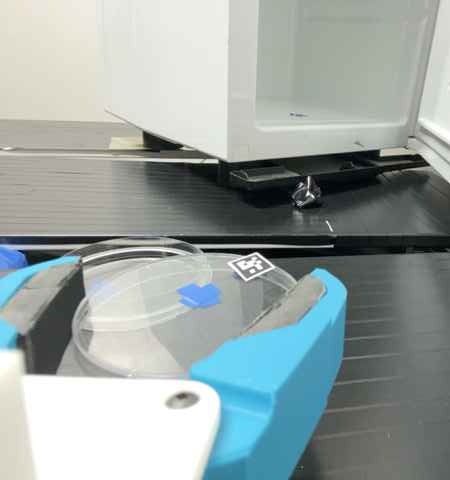}
\caption{Wrist-camera view of the lifted Petri lid held between the jaws.}
\label{fig:petri-lift}
\end{figure}

The first task was to lift a Petri dish lid, which the agent achieved
(Fig.~\ref{fig:petri-lift}; also shown in the accompanying video). The natural-language prompt was
``Lift the Petri dish lid.'' The agent explored the following procedures to
grasp the lid through Phases 1--4.

\runin{Phase 1: replay with a bias} As \texttt{AGENTS.md} suggests, the agent
considered demonstration replay with an $xyz$ bias for the displaced dish.
Difficulty estimating the bias led it to explore other methods.

\runin{Phase 2: segmentation servo} To guide motion from images of the lid, the
agent brought in SAM~3 \cite{sam3} to segment it. The controller checks
that the rim lies between level jaws before allowing closure.
It explored three visual
servo variants: a head-camera servo relative to a demonstration, a
segmentation-based servo using the left arm's camera only as an observer,
and a real-time servo tracking the segmented lid. For the spatial geometry,
it replaced a marker-plane reference with a Record3D point cloud, using
RGB-D measurements to provide depth as well as image location.

\runin{Phase 3: scene reconstruction and simulation} To reason about contact, the
agent built a multi-view RGB-D reconstruction with SAM semantics and a
MuJoCo \cite{mujoco} model of the bench, arms, and grippers. Its physics-verified
grasp search checked simulated contact and platform clearance before execution.

\runin{Phase 4: verification of the grasp} The agent calibrated gripper aperture
into empty and non-empty populations to assess contact from measured opening
rather than the close command. The jaws stayed level during descent.
The evolved program is in branch \file{wetrobo+petri}.

\subsection{Bottle cap}\label{sec:cap}

\begin{figure}[!h]
\centering
\includegraphics[width=1.00\columnwidth]{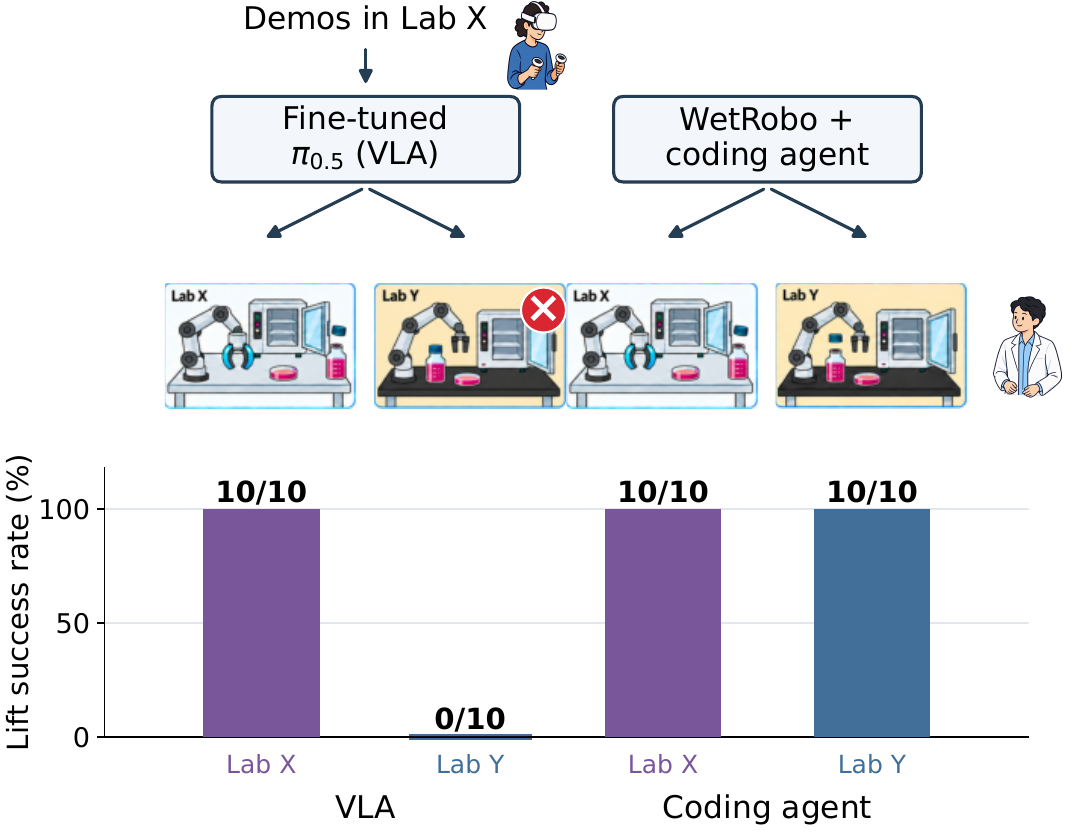}
\caption{Bottle-cap lifting: conditions (top) and success rates (bottom).
VLA rates measure evaluation after fine-tuning; coding-agent rates measure
repeated execution in the same setting after adaptation and initial task success.}
\label{fig:cap_act}
\end{figure}

The reagent bottle has a white screw cap above its wider body.
The natural-language prompt was ``Lift the bottle cap.''
This task was tested in two laboratories, Lab X and Lab Y.
Both started from the repository after the Petri trial
(Table~\ref{tab:capsettings}). Lab Y used the
Piper native gripper and Lab X an NYU string-driven Dynamixel gripper \cite{conee}. Other conditions also differed,
including arm and equipment orientation, bench color, lighting, and the
number of cameras. WetRobo with a coding agent
succeeded in both Lab Y and Lab X. In contrast, the VLA
$\pi_{0.5}$~\cite{pi05}, fine-tuned for 100k steps on 25 demonstration videos collected in Lab X,
succeeded in Lab X but failed to transfer to Lab Y
(Fig.~\ref{fig:cap_act}).
We describe Lab Y's generated files and Lab X's outcome and implementation
differences (Table~\ref{tab:capfiles}).

\begin{table}[!h]
\caption{Laboratory setting and resource use differences in Lab X and Lab Y.}
\label{tab:capsettings}
\centering
\small
\renewcommand{\arraystretch}{1.3}\setlength{\tabcolsep}{3pt}
\begin{tabular}{@{}m{0.32\columnwidth}>{\centering\arraybackslash}m{0.30\columnwidth}>{\centering\arraybackslash}m{\dimexpr0.38\columnwidth-12pt\relax}@{}}
\toprule
 & Lab X & Lab Y \\
\midrule
\multicolumn{3}{@{}l}{\emph{Laboratory}} \\
Workspace & \includegraphics[width=\linewidth]{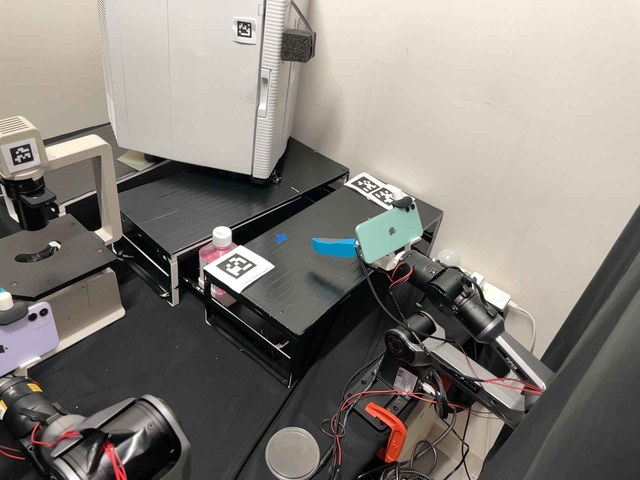}
 & \includegraphics[width=\linewidth]{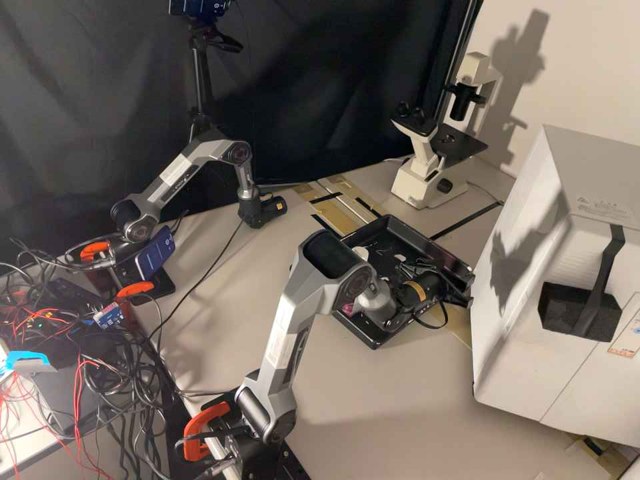} \\
Luminance (0--255) & 130.10 & 106.84 \\
Gripper & NYU & Piper native \\
Cameras & 2 (head, wrist) & 3 (head, left, on the bench) \\
\midrule
\multicolumn{3}{@{}l}{\emph{Trial}} \\
Tokens & 41.4\,M & 23.0\,M \\
Time to first grasp & 1\,h\,02\,min & 48\,min\,43\,s \\
\bottomrule
\end{tabular}
\end{table}

\subsubsection{Lab Y: the files the agent wrote}\label{sec:cap-laby}

\begin{figure*}[!t]
\centering
\includegraphics[width=\textwidth]{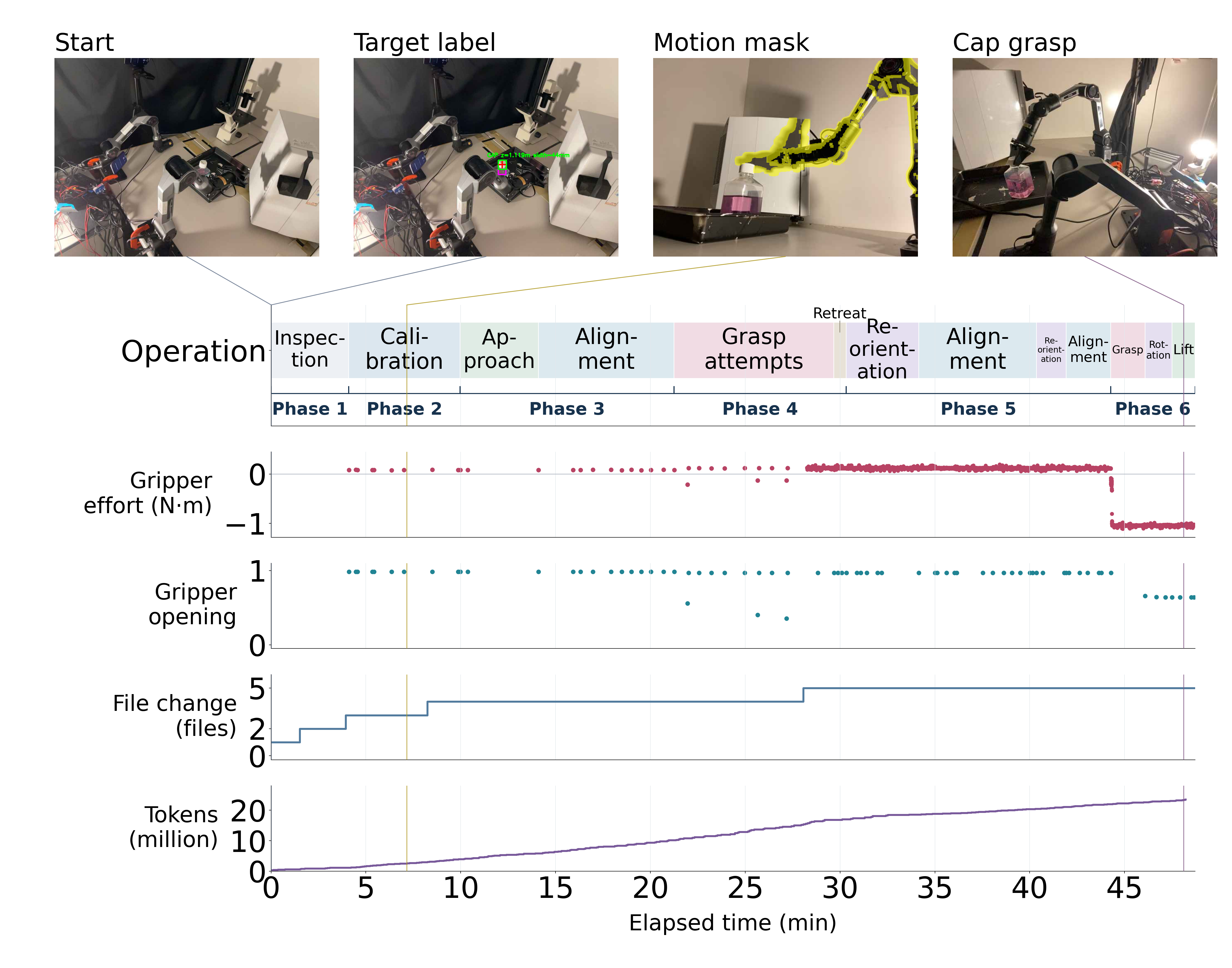}
\caption{Lab Y cap trial up to the first grasp: operations, gripper
measurements, and cumulative file and token counts.}
\label{fig:timeline}
\end{figure*}

\begin{table*}[!t]
\caption{Comparison of the evolved programs for the cap task.}
\label{tab:capfiles}
\centering\small
\renewcommand{\arraystretch}{1.3}\setlength{\tabcolsep}{8pt}
\begin{tabular}{@{}>{\raggedright\arraybackslash}p{0.15\textwidth}>{\raggedright\arraybackslash}p{0.39\textwidth}>{\raggedright\arraybackslash}p{0.39\textwidth}@{}}
\toprule
Aspect & Lab Y & Lab X \\
\midrule
Grasp evidence & measured gripper effort and opening after closure & visual
assessment and measured opening after closure \\ \addlinespace[5pt]
Motion checks & joint torque monitored during motion; stop if a limit is
exceeded & simulated collision checks before a candidate motion is executed \\ \addlinespace[5pt]
Spatial grounding & depth and intrinsics (camera frame); image-motion probes
& known-size, pre-positioned AprilTag: RGB-D to planning scene \\
\bottomrule
\end{tabular}
\end{table*}

Fig.~\ref{fig:timeline} follows the trial from the start of the session to
the first grasp (also in the accompanying video). Cap Phases 1--6 denote consecutive groups
of operations, not individual closures; phase numbering restarts for each task.

\runin{Phase 1: inspection} The agent generated
\file{src/labb_right_cap_control.py} to coordinate motion
and gripper control. Inverse kinematics (IK) converts small position changes
into joint motion with joint 6 held fixed. The program stops if measured
joint torque, the turning load on a joint, exceeds its limit.
Gripper effort, a separate load measurement, helps assess the grasp.

The agent then needed to locate the cap and move toward it in small steps
with a stopping condition. To locate the cap directly from images, it generated
\file{src/detect_culture_bottle_cap.py}, which uses OpenCV
\cite{opencv} and NumPy \cite{numpy}. A color threshold
picks out the magenta liquid and the white cap above it (``Target label''
in Fig.~\ref{fig:timeline}). Record3D depth and camera intrinsics provide
the cap's metric position.

\begin{figure*}[!t]
\centering
\includegraphics[width=\textwidth]{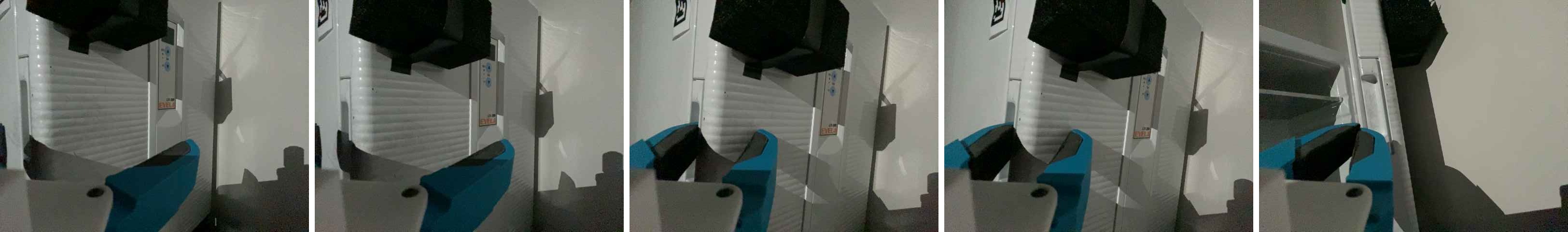}
\caption{Lab X door-opening sequence from the wrist camera. Left to right:
yaw-aligned approach, open-jaw contact, closure, proof pull, and door open.}
\label{fig:success}
\end{figure*}

\runin{Phase 2: motion probes and calibration} To relate image errors to arm motion,
the agent made small test movements and compared images
using the agent-generated \file{src/estimate_frame_motion.py}.
OpenCV's pyramidal Lucas--Kanade
optical flow \cite{lucaskanade} tracks image corners before and after
each probe. The resulting image displacements helped associate camera views
with the moving arm (``Motion mask'' in Fig.~\ref{fig:timeline}).

From three motion probes, one per
axis, the agent solved a local image Jacobian, a mapping from
small arm displacements (mm) to image displacements (pixels), for each camera.
This supplied alignment corrections. Such local
recalibration offers a mechanism for adapting to changed camera geometry,
unlike direct transfer of the fine-tuned VLA evaluated here.

\runin{Phase 3: approach and alignment} With this mapping, the arm approached the cap in
a sequence of position steps, with head and other camera images read between
steps. Contact was read from the images rather than the torques: when the
bottle moved in the image while the torque rise stayed within its limit,
the agent backed the arm half a step along the path it had come.

\runin{Phase 4: grasp attempts and retreat} The images suggested the jaws were
ready, so the agent commanded closure. Three closures failed: the aperture
(measured jaw opening) followed the command, effort remained at the empty-close level,
and images showed the jaws above or beside the cap. The arm retreated.

\runin{Phase 5: reorientation and alignment} To correct the missed grasps,
the agent turned the arm with joint commands and brought it back to the cap, adding
a wrist-hold variant to the same execution program, in which only the first three
joints are active, so the wrist stays still while the jaws are centered.

\runin{Phase 6: grasp, rotation, and lift} This time the opening stopped at 0.68
against a command of 0.50, and the effort read $-1.05$\,N$\cdot$m against
$-0.1$ to $-0.2$\,N$\cdot$m for an empty close: the jaws held the cap.
The agent added a twist motion to that execution program, maintaining the fingertip position while
rotating about the world $z$ axis. After three twists, a short lift
confirmed the grip---the first grasp (``Cap grasp'' in Fig.~\ref{fig:timeline}).
Lab Y's evolved program is in branch \file{wetrobo+petri+cap@labY}.

\subsubsection{Lab X: how the code evolved differently than in Lab Y}\label{sec:cap2}

Differences in grippers, available feedback, camera arrangements, and bottle
placement shaped the agent's choices across laboratories (Tables~\ref{tab:capsettings}
and \ref{tab:capfiles}). Lab X's NYU gripper lacked effort feedback. For motion,
Lab X's agent-generated \file{src/run_culture_media_cap_grasp.py} used
MuJoCo \cite{mujoco} collision checks before moving, whereas Lab Y's
agent-generated \file{src/labb_right_cap_control.py} stopped on excessive
joint torque. The Lab X runner reused the simulated-contact helper
generated during the Petri task in \file{src/run_codexless_thin_object_grasp.py}.
The scene represented the reagent bottle recessed between supports rather
than on a tabletop; the runner refreshed it using Lab X's
pre-positioned AprilTag. For grasp assessment, Lab X combined images with
opening: the agent-generated
\file{rollout/media_cap_target.py} checked for the cap between closed jaws,
and the runner checked support movement and opening, whereas Lab Y used
effort and opening. The task also succeeded
in Lab X. From the same kit, the programs evolved differently to suit each
laboratory's conditions.
Lab X's evolved program is in branch \file{wetrobo+petri+cap}.

\subsection{Incubator door}\label{sec:door}

The incubator handle is a horizontal door recess accepting fingers or
gripper jaws for pulling. The natural-language prompt was
``Open the incubator door.'' Execution required approach, jaw
alignment, grasping, pulling, and opening verification (Fig.~\ref{fig:success})
in the same laboratories as the cap.

\subsubsection{Lab Y}

Unreliable head-camera alignment prompted a pre-contact stop and return
home, leaving the door closed. Lab Y's snapshot is branch
\file{wetrobo+petri+cap+door@labY}.

\subsubsection{Lab X}

The task was achieved by progressively adding image-based alignment,
grasp checks, and door-state estimation to the evolved program
(Fig.~\ref{fig:success}; also in the accompanying video).

\begin{figure}[!t]
\centering
\includegraphics[width=1.00\columnwidth]{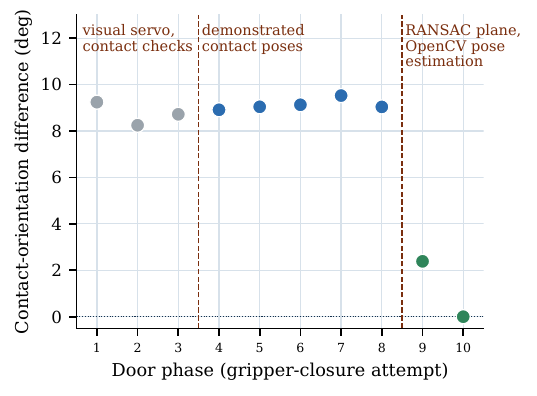}
\caption{Pre-close gripper orientation across recorded Lab X door
attempts, relative to the successful final attempt.
}
\label{fig:contact}
\end{figure}

The agent generated \file{src/run_incubator_door_demo.py} to execute motion
stages, then revised it and added analysis modules as alignment and contact
problems emerged. Fig.~\ref{fig:contact} maps tool additions onto the
pre-close poses---the gripper's position and orientation just before closing---
recorded at each attempt, showing the orientation difference from the final
successful attempt.

\begin{figure}[!tb]
\centering
\includegraphics[width=1.00\columnwidth]{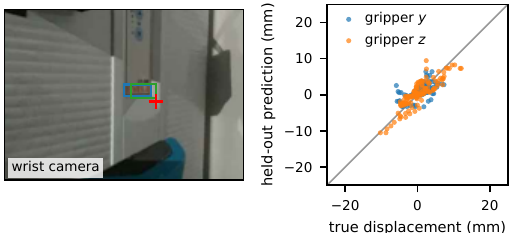}
\caption{Example of Codex's analysis: learning jaw alignment to the handle.
A regression model maps wrist-camera label-feature residuals to end-effector
displacement toward the demonstrated closing pose. Left: detections before
(blue) and after (green) lateral correction, with the goal in red. Right:
held-out predicted versus true lateral displacement in the gripper frame.
Redrawn from saved trial data and an offline regression evaluation.}
\label{fig:regression}
\end{figure}

\begin{figure}[!b]
\centering
\includegraphics[width=1.00\columnwidth]{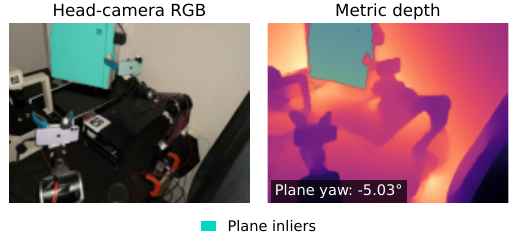}
\caption{Example of Codex's analysis: door orientation from OpenCV
tag-based pose estimation and a NumPy RANSAC plane fit. Overlays show
plane inliers; yaw is estimated in the robot frame.
Rendered by rerunning the evolved program on saved trial RGB-D data.}
\label{fig:rgbd}
\end{figure}

\runin{Phases 1--3: measuring alignment and contact}
The agent selected the manufacturer's small red label on the door as a
visual alignment cue. Fixed to the same rigid door as the handle, it
remained visible when the gripper partly hid the recess. The agent generated
\file{rollout/incubator_door_visual.py} to measure label features with OpenCV
and fit a ridge-regression model with NumPy. With the agent-generated
\file{src/compile_incubator_door_demos.py}, this model learned from
demonstrations how image-feature differences predict the gripper displacement
needed to reach the demonstrated closing pose (Fig.~\ref{fig:regression}).
Nevertheless, alignment alone was insufficient:
Phases 1--3 failed the check for a stable non-empty aperture, meaning a jaw
opening that remains wider than an empty closure. Measuring aperture
distinguished commanding closure from holding the handle.

\runin{Phases 5--8: recovering contact and checking retention}
During Phases 5--8, the agent revised the motion runner's open-jaw approach,
release with the wrist held still, and grasp checks. These were control-code changes rather
than new external-tool additions; the orientation difference in
Fig.~\ref{fig:contact} remained broadly unchanged. Aperture distinguished
empty closure, near 0.005 of full opening, from a handle hold at 0.3--0.4.
The runner tested the hold with a 5\,mm proof pull---a short pull before
continuing to open the door. Slipping during
subsequent pulling motivated slower pull segments with aperture checks
between them, followed by stopping, retreating, and opening when the hold
was lost. This made loss of contact observable, but did not resolve the
remaining approach-orientation mismatch.

\runin{Phase 9: replacing shadows with measured door geometry}
Then, the agent realized that shadows gave an unreliable cue to the door's angle. The agent generated
\file{rollout/incubator_door_plane.py} to measure it from depth instead.
Pose estimation from visual tags using the external OpenCV library \cite{opencv} expressed
depth points in robot coordinates. A NumPy RANSAC fit \cite{ransac} then
identified the door plane while rejecting inconsistent points
(Fig.~\ref{fig:rgbd}). Its yaw, the rotation about the vertical axis, was
measured against a saved observation of the closed door. The agent edited
the motion runner to rotate the approach by this angle. This helped the
robot achieve a much better gripper angle.

\runin{Phase 10: integrating motion with endpoint verification}
Before proceeding to open the door, the agent sought to strengthen the
criteria for determining whether it was open or closed, and therefore
generated \file{src/run_incubator_door_autonomy.py} and
\file{rollout/articulated_appliance.py}. The former coordinated the motion
runner and analysis modules: measured door yaw corrected the demonstrated
pre-close orientation, and the wrist-image regression supplied a lateral
position correction before contact, closure, and pulling.
The latter used OpenCV image registration to align head-camera
RGB-D observations with stored open and closed references, then compared
the door-region depth. Its relative error is the depth mismatch to a
reference divided by the depth separation between the open and closed references. The recorded
execution reached the open state (Fig.~\ref{fig:success}); relative error
against the open reference changed from 1.417 to 0.179.
It was therefore recorded as successfully open.

\begin{figure}[t]
\centering
\includegraphics[width=\columnwidth]{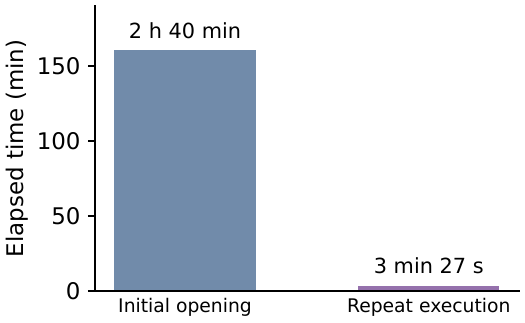}
\caption{Time to door opening}
\label{fig:door_replay_time}
\end{figure}

Once accomplished, the door-opening task was repeated in the same setting
in 3\,min 27\,s, including parser repairs and restarts
(Fig.~\ref{fig:door_replay_time}).
Lab X's evolved program, the kit's code exemplar, is in branch
\file{wetrobo+petri+cap+door}.

\suppressfloats[t]

\section{Discussion}\label{discussion}

\runin{Coding agents are good robot controllers} The coding agent succeeded
in all three tasks examined here, which represent basic operations in wet
laboratories. Their relevance across laboratory procedures suggests that
coding agents may eventually support full-length cell-culture workflows and
more demanding tasks; such long-horizon performance remains to be evaluated.

\runin{Code evolution} We observed the coding agent autonomously evolving
code during trials. External tools, including OpenCV pose estimation and
image registration, NumPy RANSAC plane fitting, SAM segmentation, and MuJoCo
collision checks, helped it adapt. These observations support allowing the
agent to obtain tools as needed rather than prescribing a fixed tool set.
Petri-task implementations supported later tasks: the cap program reused
the agent-generated \file{src/run_codexless_thin_object_grasp.py} for Cartesian
motion, joint trajectories, and contact checks. The door program reused the
live-camera class in the agent-generated \file{src/run_demo_relative_servo.py},
which wraps camera capture. Reuse shortened execution in the same setting
(Fig.~\ref{fig:door_replay_time}). The evolved programs are available as
exemplar branches. Although the present comparison starts from the WetRobo
kit in Lab X and Lab Y, adapting the exemplar evolved in Lab Y to Lab X
may further reduce adaptation cost; this remains a question for future studies.

\runin{WetRobo + coding agents are more flexible than VLAs} In the cap
comparison, the VLA fine-tuned on Lab X demonstrations succeeded there but
failed to transfer to Lab Y, whereas the coding agent succeeded in both.
This comparison tests cross-laboratory VLA transfer against on-site code
adaptation. For biological experimentalists, this offers a practical route
to deployment without collecting local teleoperation data or training a
neural network for each laboratory.

\runin{Human effort and cost} Delegating adaptation can
reduce the need for local teleoperation, but incurs LLM token consumption
(Table~\ref{tab:capsettings}) and associated energy demand. Agent and kit
improvements should reduce these costs. A comparison of total resource use
should also include VLA fine-tuning costs.

\runin{Sharing AGENTS.md across laboratories} Sharing and refining
\texttt{AGENTS.md} across laboratories is expected to improve robot performance
and reduce adaptation costs as procedural knowledge accumulates.
Recorded actions already inform these rules, including additions from the
Lab Y cap trial. Natural-language task descriptions or
egocentric video may eventually replace our teleoperation recordings;
passive human-work videos could yield skill rules for robot validation.
Shared instructions could propagate improvements without the per-site
retraining of separate VLA policies.

\runin{Limitations} The trials used one coding agent and one model, leaving
performance with other agents and models untested. Task outcomes came from
individual adaptation trials with successive attempts, rather than repeated
independent adaptations. The coding-agent success rates in
Fig.~\ref{fig:cap_act} describe repeated executions in the same setting after
adaptation and initial success. Camera placement
and installed reference objects also condition the reported outcomes.
The Petri result concerns lid lifting; lifting and transporting the dish
itself remains an untested challenge. Slipping observed after a verified lid
lift motivates future work on sustained grasp retention for dish transport.
Broader equipment, task, and laboratory coverage is needed to assess generality.

\section{Conclusion}\label{conclusion}

Reproducible laboratory robots must accommodate environmental variation
that can impair VLA policies. We provide WetRobo so that biological experimentalists
can assemble the specified equipment, give natural-language tasks, and
delegate adaptation to a coding agent without performing teleoperation or
neural-network training. Codex produced evolved programs and action records
for Petri-lid lifting, door opening, and cap removal, all in real-world laboratories. The cap task was reproduced across laboratories Lab X and Lab Y
despite changes in gripper and camera placement. A VLA fine-tuned on
demonstrations collected in Lab X succeeded there but failed to transfer to Lab Y, whereas
the coding agent succeeded in both. These results establish on-site code
adaptation as a practical route to biological laboratory automation.
Incorporating the resulting rules into \texttt{AGENTS.md} makes each
installation a source of reusable knowledge. Sharing this knowledge across
laboratories is expected to support less costly, faster, and safer wet-lab
robots across different arms, grippers, and coding agents.

\section*{Acknowledgment}
We thank Naruki Yoshikawa, Kazuki Takahashi, Liyao Wang, Xiaotian Xue, and Kaoru Shibutani for their help with the robot.
This work was supported by UTokyo-Google AI Symbiotic Future Society Program.
OpenAI Codex assisted with drafting, editing, and proofreading the manuscript.
Generative image tools accessed through Codex were used to create and edit
the illustrations in Figs.~\ref{fig:concept}, \ref{fig:skills}, \ref{fig:agent-control-flow},
and \ref{fig:cap_act}.

\end{document}